\documentclass[sigconf,nonacm]{acmart}

\usepackage{cleveref}
\usepackage{nicefrac}

\usepackage{multirow}
\usepackage{makecell}
\usepackage{subcaption}
\usepackage{adjustbox}
\newcommand{\showarticletitle}[1]{#1}

\newcommand{\dt}[1]{\multicolumn{2}{c}{#1}}
\newcommand{\ci}[1]{\,$\pm$\,#1}

\renewcommand\footnotetextcopyrightpermission[1]{}
\begin{document}

\title{Joint user--item allocation for e-commerce \\ marketing campaigns}

\author{Maja Lindstr\"{o}m}
\affiliation{%
  \institution{Integrated Science Lab\\ Department of Computing Science}
  \city{Ume\aa{} University}
  \country{Sweden\\}
  \institution{Siftlab AB}
}
\email{maja.lindstrom@umu.se}

\author{Natalija Glisovic}
\affiliation{%
  \institution{KTH Royal Institute of Technology \\ IKEA Retail (Ingka Group)}
  \city{Stockholm}
  \country{Sweden}}
\email{glisovic@kth.se}

\author{Jan von Pichowski}
\affiliation{%
  \institution{Chair of Machine Learning for Complex Networks\\ Center for Artificial Intelligence and Data Science (CAIDAS)}
  \city{University of W\"{u}rzburg}
  \country{Germany}
}

\author{Tommy L\"{o}fstedt}
\affiliation{%
 \institution{Department of Computing Science}
 \city{Ume\aa{} University}
 \country{Sweden}}

\author{Martin Rosvall}
\affiliation{%
  \institution{Integrated Science Lab \\Department of Physics}
  \city{Ume\aa{} University}
  \country{Sweden\\}
  \institution{Siftlab AB}}

\begin{abstract}
    Web platforms run many marketing campaigns at once, and for each one they must decide which products to promote and which users to target. The best users depend on the products, and the best products on the users. The campaigns also compete with one another, because each user and each item may belong to only one of them and each audience is capped by the campaign budget. Still, existing approaches fix one side and search the other, choosing an assortment and then looking for its audience, or ranking items for one user at a time. Either way they solve a restricted version of the full problem, implying that a jointly optimized assignment would be at least as good and possibly far better. What the restriction costs, and whether the joint problem is solvable at the scale campaigns are run, remain unknown. Here we formalize the joint task as auto-targeting: assigning users and items to multiple disjoint, capacity-bounded campaigns that maximize the affinity between them. We adapt constrained spectral biclustering, greedy local search with pairwise swaps, and a multi-armed bandit formulation to the task and compare them against the conventional decoupled baseline. On synthetic data, two Amazon Reviews benchmarks, and two proprietary commercial datasets, auto-targeting raises campaign lift over the best decoupled pipeline, from around 20\% on the two smaller, denser datasets to more than threefold on the two largest.
    Biclustering yields the highest campaign quality across datasets, while bandit-based methods run fastest on the largest ones.
\end{abstract}


\begin{teaserfigure}
\centering
\begin{subfigure}{.32\linewidth}
  \centering
  \includegraphics[trim=5px 5px 5px 30px, clip, width=\linewidth]{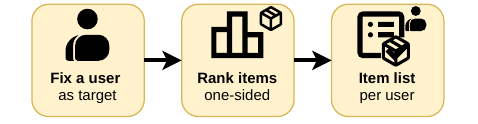}
  \caption{Traditional recommender systems {\normalfont fix one user at a time and rank items for that user. Users are not assigned into disjoint groups, and the same items can surface for every user.}}
  \label{fig:sub1}
\end{subfigure}%
\hfill
\begin{subfigure}{.32\linewidth}
  \centering
  \includegraphics[trim=5px 5px 5px 30px, clip, width=\linewidth]{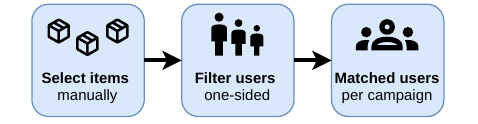}
  \caption{Traditional campaign design {\normalfont fixes an item assortment first, such as a product category, and then searches for the audience most attracted to it, so items are never selected with respect to users.}}
  \label{fig:sub2}
\end{subfigure}%
\hfill
\begin{subfigure}{.32\linewidth}
  \centering
  \includegraphics[trim=5px 5px 5px 30px, clip, width=\linewidth]{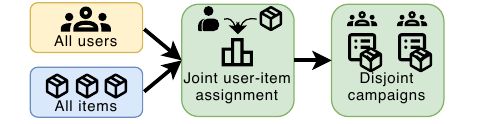}
  \caption{Auto-targeting{\normalfont{,} introduced here, jointly assigns users and items to disjoint campaigns, so user and item assignments are coupled throughout the process rather than optimized one at a time.}}
  \label{fig:sub3}
\end{subfigure}
\caption{The problem of auto-targeting fills the gap between recommender systems and campaign design: both existing paradigms decouple one side of the assignment from the other, while auto-targeting optimizes users and items jointly.}
\label{fig:test}
\end{teaserfigure}

\maketitle

\section{Introduction}

Matching users to items is central to personalization on the Web. When retailers run promotional campaigns through advertising, personalized promotions, or email marketing, they face a combinatorial assignment problem. The goal is to assign users and items to campaigns such that each campaign contains a cohesive cohort, while ensuring that each user and item appears in only one campaign~\cite{kim2009Multicampaign}. Existing approaches decouple this joint problem, fixing either the audience and ranking items for each user~\cite{zhou2023comprehensive} or the item set and searching for its audience afterwards~\cite{sedlarova2025robust, rahman2024linknet} (Figure \ref{fig:test}). 
This decoupling breaks a bidirectional dependency where the best users for a campaign depend on which items it contains, and the best items depend on who will be interested in them. Searching one side while holding the other fixed misses groupings that emerge only from joint optimization over interaction patterns.
Since every decoupled assignment is a special case of the joint problem, the joint optimum is at least as good.

The constraints come from how campaigns are delivered. A channel such as promotional email or onsite advertisement has a finite number of slots, and each campaign has a budget that caps the audience it can reach, so each campaign's size is bounded in both items and users. Placing a user in several campaigns causes marketing fatigue, so campaigns must be disjoint. A campaign assignment therefore selects disjoint, size-bounded groups of users and items. Recommender systems rank items for one user at a time and never produce such an assignment, but they hold the input for one. In modern e-commerce systems, users and items are represented as vector embeddings learned from historical interactions, such as purchases or clicks~\cite{qu2024budgeted}. Proximity in this shared space reflects affinity~\cite{salha2023consistency}. We take these embeddings as given and build the assignment from them. Building the assignment is a combinatorial search over hundreds of thousands of users and items, so a method that is accurate but slow is impractical. Three questions are open: what the decoupling costs, whether the joint problem is tractable, and which method builds the best assignment.

We propose a joint campaign assignment task, which we refer to as \emph{auto-targeting}, that assigns users and items to disjoint, size-bounded campaigns that maximize within-campaign affinity. The campaign structure thus emerges from the affinities instead of being fixed in advance. To solve the assignment problem, we adapt three methods that differ in what they reason over. Biclustering identifies dense blocks of the user--item affinity matrix, whereas alternating greedy optimization with local swap refinement and a multi-armed bandit formulation score individual users and items. Simulated annealing serves as a baseline, and we compare all four against the decoupled pipeline used in practice.

We make three main contributions.
\begin{itemize}
    \item We formalize auto-targeting as a joint optimization problem over users, items, and campaign structure, a task set by the constraints of Web campaign delivery.
    
    \item We quantify what the coupling is worth. We compare against the decoupled pipeline used in practice, solving its user-assignment stage exactly so that any gap is due to the item selection. Since every decoupled solution is feasible for our formulation, the gaps are lower bounds. The best joint method exceeds the best decoupled pipeline by 17\% and 21\% on the two smaller, denser datasets, and by 220\% and 380\% on the two largest.

    \item We adapt constrained spectral biclustering, greedy local search, and multi-armed bandits to the task and evaluate them on a synthetic benchmark, two Amazon Reviews datasets, and large-scale proprietary commercial data. Biclustering, which reasons over blocks of the affinity matrix, matches or beats the methods that score individual users and items on every dataset and wins clearly on the two largest, while bandit-based methods run fastest there.
\end{itemize}

\section{Related Work}
Campaign assignment in industry is commonly a two-stage pipeline: a retailer decides what to promote and then who should see it. Work on client selection \cite{nobibon2011Optimization,  gunnarsson2019optimizing}, on budget allocation across campaigns and channels \cite{sedlarova2025robust, eshghi2020spread, han2020contextual}, and on audience identification for a fixed assortment \cite{rahman2024linknet} all treat the item set or the audience as fixed, and optimize the remaining side. The multicampaign assignment problem~\cite{kim2009Multicampaign} comes closest to ours: it assigns customers to several concurrent campaigns under capacity limits, but with the content of every campaign given in advance. When products are fixed in advance, assigning users is typically handled with graph-based audience techniques~\cite{rahman2024linknet, zhuang2020hubble} or bipartite matching~\cite{mehta2013adbook, charles2010fast}. More recent work combines offline optimization with real-time pacing, planning campaign allocations in advance and adjusting them during delivery to meet budget and exposure targets \cite{zhao2020jointly, zhao2023unimatch}. Still, these approaches decouple item selection from user assignment when campaigns are formed: they learn a joint representation but assign one side at a time. Across all these lines of work, either the campaign structure is predefined or one side of the user--item pairing is fixed while the other is optimized. Auto-targeting fixes neither. Only the number of campaigns and their size bounds are given in advance, while which items each campaign contains and which users receive them are decided together, in one optimization problem.

We draw on three lines of algorithmic work.
Spectral co-clustering partitions the rows and columns of an affinity matrix jointly through a singular value decomposition relaxation of a bipartite graph cut followed by $k$-means rounding~\cite{dhillon2001co, zha2001bipartite, dhillon2004kernel, wu2016general, su2024spectral}. These methods recover block structure efficiently, but no variant bounds the size of a block or keeps blocks from overlapping.
Greedy submodular maximization~\cite{nemhauser1978analysis, leskovec2007cost, krause2014submodular} and local search with pairwise swaps~\cite{kernighan1970efficient, kanungo2004local, awasthi2014local} are standard tools for subset selection, with guarantees that assume a single fixed set to select from. Here the value of a user depends on the items in the campaign and the value of an item on the users who receive it, so there is no fixed set and the guarantees do not carry over.
Multi-armed bandits such as UCB1~\cite{auer2002ucb} and Thompson Sampling~\cite{thompson1933likelihood} balance exploration and exploitation over independent arms with stationary rewards, and extensions handle rewards that change over time~\cite{chen2023nonstationary, qi2025thompson} or arms that share a capacity budget~\cite{agrawal2014bandits, liu2022nonstationary}. Here each campaign's reward shifts as the opposing side is reassigned, and all campaigns draw on the same users and items, so the arms are neither independent nor stationary, and the drift is caused by the assignment itself. None of the three methods handles the coupling between users and items that defines auto-targeting. We therefore adapt each to the task (Section~\ref{sec:approaches}). We oversample candidate biclusters, prune them to the size bounds, and keep the best non-overlapping ones; we alternate greedy allocation between users and items and refine it with capacity-aware swaps; and we score bandit arms against the current assignment of the opposing side, with discounted reward statistics to track the drift.

\section{Auto-Targeting}\label{sec:method}

We define auto-targeting as the problem of jointly assigning users and items to multiple disjoint, size-bounded campaigns to maximize within-campaign affinity, the proximity of users and items in a shared embedding space. Neither the campaign structure nor either side of the assignment is fixed in advance, and at Web scale the search space is vast. Our datasets span up to 500,000 users and 149,515 items. Auto-targeting is a selection problem, not a recommendation problem, since ranking items for each user would place the same popular items in every campaign. Recommendation instead supplies the input: any engine that embeds users and items in a shared space, so that proximity reflects affinity, can provide it, and our methods make no assumption about how the embeddings were found.

\subsection{Problem Formulation}

Let $K$ denote the number of campaigns, and for each $k \in \{1, \dots, K\}$, let $I_k$ and $U_k$ denote the sets of assigned items and users, respectively. We then propose the problem
\begin{align} \label{eq:problem}
    \max_{\{I_k, U_k\}_{k=1}^K}
        & \sum_{k=1}^{K} \sum_{i \in I_k} \sum_{u \in U_k} \mathrm{affinity}(i, u) \\
    \mathrm{subject~to~~}
        & I_k \cap I_j = \varnothing,\;                                 \forall k \neq j, \nonumber\\
        & U_k \cap U_j = \varnothing,\;                                 \forall k \neq j, \nonumber\\
        & \tau^{I}_{\min} \leq |I_k| \leq \tau^{I}_{\max},\;            \forall k, \nonumber\\
        & \tau^{U}_{\min} \leq |U_k| \leq \tau^{U}_{\max},\;            \forall k, \nonumber
\end{align}
where $\tau^{I}_{\min}, \tau^{I}_{\max}$ and $\tau^{U}_{\min}, \tau^{U}_{\max}$ bound the
number of items and users per campaign. The first two constraints ensure that each item
and each user is assigned to at most one campaign, and the last two bound each campaign's size on both sides. In the experiments here, we let $\tau^{I}_{\min} = \tau^{I}_{\max} = I$, so that every campaign
contained exactly $I$ items, while leaving the number of users to vary within its bounds. This reflects common practice, where campaigns use a fixed product assortment, while the size of the target audience can vary depending on budget, channel capacity, and delivery constraints.

The constraints in Equation~\eqref{eq:problem} are those of a deployed commercial campaign workflow, and they form a base case rather than a limit of the formulation. Each can be relaxed by changing a single constraint while the objective and the coupling between users and items stay unbroken. Allowing a user to appear in a bounded number of campaigns, rather than exactly one, expresses frequency capping, and lifting the same restriction on the item side lets a product feature in several campaigns at once. Per-campaign budgets replace the bound on audience size with a bound on total cost, and eligibility rules, by region or stock, forbid individual user--item pairs. We evaluate the base formulation throughout, since it is the setting our data comes from.

\subsection{Combined User--Item Embedding Space}

While the formulation in Equation~\eqref{eq:problem} does not depend on a specific embedding, the affinity score is derived from alignment in a given joint embedding space, where higher affinity corresponds to closer alignment between users and items.  Users, $u$, and items, $i$, are represented in a shared $D$-dimensional space by the rows
$\mathbf{U}_u \in \mathbb{R}^{D}$ and $\mathbf{V}_i \in \mathbb{R}^{D}$ of the embedding matrices $\mathbf{U} \in \mathbb{R}^{N_u \times D}$ and
$\mathbf{V} \in \mathbb{R}^{N_i \times D}$, so that users and items forming coherent
clusters in this space are expected to be assigned to the same campaign.

For the real-world datasets, we adopt embeddings from an inter\-action-driven recommendation framework~\cite{theo}: each item $i$ has a learnable embedding $\mathbf{V}_i$, while user embeddings
$\mathbf{U}_u$ are derived by aggregating and normalizing the embeddings of recently
interacted items, yielding a contextual representation of current preferences. To address sparsity and cold-start effects, visual and textual item features are combined via a vision--language model into semantic meta-embeddings, used to initialize or refine item embeddings when interaction data is limited. For the synthetic data, embeddings are constructed directly from a known ground-truth campaign structure: users and items within the same campaign
are sampled from Gaussians centered on campaign-specific latent vectors, giving high within-campaign affinity and controlled separation between campaigns.

\subsection{Affinity Values of User--Item Combinations}

Equation~\eqref{eq:problem} is described in terms of a generic affinity function. Any function will do, provided it is non-negative and increases with alignment in the embedding space. Non-negativity keeps the size bounds meaningful: the objective sums over assigned pairs, so negative entries would shrink every campaign to its smallest allowed size regardless of content. We use $g(x) = \exp(x)$, making $\mathbf{A}_{iu} = \exp(\langle \mathbf{V}_i, \mathbf{U}_u \rangle)$ the affinity between item $i$ and user $u$, with $\mathbf{A} \in \mathbb{R}^{N_i \times N_u}_{\geq 0}$. The inner product captures the similarity between embeddings, and the exponential ensures positivity while increasing the contrast between high- and low-affinity pairs. Because the objective sums affinities rather than ranking them, the transform affects which campaigns are optimal, not just their scores. The exponential favors a few very strong pairs over many moderate ones. The resulting affinity matrix, $\mathbf{A}$, defines a weighted bipartite graph between items and users and serves as the basis for the subsequent optimization methods.

\subsection{Affinity-Based Evaluation Metrics}

Total affinity is the summed affinity over all assigned user--item pairs. Alone, it can be misleading since it rewards a method for assigning more users, it takes a different scale on every dataset, and it hides how the affinity is divided between users. We therefore use three additional evaluation metrics, each addressing one of these problems.

Utility measures the total affinity across all campaigns, consistent with prior work that frames recommendation as maximizing summed user-side utility \cite{usunier2022utility},
\begin{equation}
    \text{Utility} = \sum_{k=1}^{K} \sum_{u \in U_k} \sum_{i \in I_k} A_{iu}.
\end{equation}
Utility corresponds directly to the objective in Equation~\eqref{eq:problem}. However, it depends on the number of assigned users and items, and may favor methods that assign more users within the allowed constraints.
To account for the number of users and items, quality (the average affinity) is evaluated within each campaign,
\begin{equation}
    \text{Quality}_k = \frac{1}{|U_k|\,|I_k|} \sum_{u \in U_k} \sum_{i \in I_k} A_{iu},
\end{equation}
and captures the coherence of user--item assignments independent of the campaign size.

Lift normalizes quality relative to the global average affinity,
\begin{equation}
    \text{Lift}_k = \frac{\text{Quality}_k}{\frac{1}{N_u N_i} \sum_{i=1}^{N_i} \sum_{u=1}^{N_u} A_{iu}},
\end{equation}
and measures the improvement over a random assignment \cite{perlich2014lift}. A lift greater than 1 indicates above-chance affinity.

Finally, the Gini coefficient \cite{do2022gini} measures how utility is distributed across users. Let $v_u$ denote the average affinity of user $u$ to the items in their assigned campaign. The Gini coefficient is
\begin{equation}
    G = \frac{\sum_{u=1}^{N_u} \sum_{v=1}^{N_u} |v_u - v_v|}{2 N_u \sum_{u=1}^{N_u} v_u}.
\end{equation}
A value of 0 indicates uniform utility across users, while higher values indicate that a small subset of users accounts for most of the total affinity.
All four metrics are computed from the same affinity matrix the methods optimize over, so together they measure how effectively the objective was realized and not whether the resulting campaigns are meaningful in themselves. We complement these metrics with a qualitative inspection of the selected assortments in Section~\ref{sec:analysis}.

\section{Approaches for Auto-Targeting}\label{sec:approaches} 
Our contribution is the task, not the solvers. We adapt three standard families of algorithms to auto-targeting, chosen because they differ in what they reason over. Spectral biclustering functions on blocks of the affinity matrix, while greedy allocation and bandit optimization score individual users and items. Simulated annealing works as a baseline and moves one entity at a time within one campaign at a time. All four use the same affinity matrix and the same constraints, so any difference between them is attributable to the allocation strategy alone, and all are reference points for later work to improve on. We describe below what we changed in each. The full algorithmic detail is given in Appendix~\ref{app:implementation}. 

\paragraph{Constrained spectral biclustering.} Spectral co-clustering~\cite{dhillon2001co} partitions rows and columns of the affinity matrix jointly but enforces neither size limits nor disjoint coverage, so its output is not a feasible campaign assignment. We generate $\alpha K$ candidate biclusters instead of $K$, where $\alpha \in \{2, 3, 4, 5\}$. We then prune each to the size bounds by keeping the items and then the users with highest average affinity to the opposing side, discard candidates that fall below the minimum size, and select the $K$ densest non-overlapping survivors. Unlike the other methods it never evaluates Equation~\eqref{eq:problem} directly, and optimizes a normalized-cut relaxation instead. 

\paragraph{Greedy local search.} Greedy allocation assumes one side of the assignment is fixed. We alternate it between users and items, tracking per-campaign capacity so that entities which cannot be placed without violating a constraint are left unassigned, and follow each pass with pairwise swaps between campaigns, accepted when they improve the objective and preserve feasibility. 

\paragraph{Bandit optimization.} Classical bandit algorithms assume arms are independent and their rewards stationary. Neither holds here. A campaign's reward depends on which entities are currently assigned to the opposing side, and all campaigns draw on the same capacity. We score each entity by its affinity to the campaign's current members on the opposite side, add a UCB1 or Thompson bonus scaled to the magnitude of that reward, assign entities in descending order subject to capacity, and alternate user and item phases. Reward statistics are discounted exponentially so that they track the shifting composition instead of averaging over it. 

\paragraph{Simulated annealing baseline.} As a baseline we decompose the problem into $K$ subproblems solved in sequence, each optimizing one campaign over the users and items still unassigned, using add, remove, and swap moves under a Metropolis acceptance criterion. This is the only method that optimizes campaigns one at a time, and it is therefore very slow as can be seen in the results.

\section{Decoupled baselines}
The methods above optimize items and users \emph{jointly}. To measure the value of this
joint optimization, we compare them against the decoupled pipeline used in practice, where the items are fixed first --- by product category, seasonal theme, or popularity ranking --- and the audience is found afterwards for the items already chosen. Each baseline operates in two stages: (i) select $I$ items for each campaign using heuristics that have no access to the user assignment, and (ii) freeze the selected items and assign users to campaigns.

We solve the second stage exactly, so that any gap to the joint methods is due to the item selection alone. Once the items are fixed, the objective in Equation~\eqref{eq:problem} is linear in the user assignment. Let
$s^u_k = \sum_{i \in I_k} \mathbf{A}_{iu}$ denote the affinity of user $u$ to campaign $k$ (Appendix~\ref{app:score_aggregation}). The objective can then be written as $\sum_u s^u_{k(u)}$, where $k(u)$ is the campaign to which user $u$ is assigned. With the bounds $\tau^{U}_{\min} \leq |U_k| \leq \tau^{U}_{\max}$, this assignment is a transportation problem. Its constraint matrix is unimodular, so the linear relaxation has an integral optimum and the resulting linear program gives the optimal integer assignment. We limit the program to the $K\tau^{U}_{\max}$ highest-scoring users for each campaign. This does not change the optimum, since at most $K\tau^{U}_{\max}$ users can be assigned in total. Therefore, any assigned user outside a campaign's shortlist can be replaced by an unassigned shortlisted user without decreasing the objective.

Any gap between a baseline and the joint methods is due to the item selection, since the user assignment is optimized exactly in both cases. Because every decoupled solution is also feasible for Equation~\eqref{eq:problem}, the joint
optimum is at least as large as any solution these pipelines produce. The reported gaps
therefore lower-bound what the joint optimum gains over the decoupled solutions we construct. 

We select items using four heuristics that cover high average affinity, shared audience, both, and neither. \texttt{random} draws $KI$ items uniformly at random and splits them into $K$ groups of $I$, exploiting neither property. \texttt{popular} selects the top $KI$ items by average affinity across users and assigns them to campaigns in round-robin order. \texttt{kmeans} clusters the item embeddings $\mathbf{V}_i$ into $K$ groups and selects the $I$ items closest to each centroid, promoting coherence without considering affinity. \texttt{top\_dim} selects the $I$ top items along each of the $K$ highest-variance embedding dimensions, capturing both properties in part. All four heuristics use the same $K$, $I$, and audience bounds as the joint methods, with items disjoint across campaigns. \texttt{random} and \texttt{kmeans} depend on the random seed and are therefore repeated over the same 10 seeds as the other methods. \texttt{popular} and \texttt{top\_dim} are deterministic and are run once.

\section{Experimental Setup}

\begin{table*}[t]
    \centering
    \caption{Dataset characteristics. Users and items refer to counts after
    preprocessing. The last column gives the allowed range of users per campaign.
    All datasets use $K=5$ campaigns and $I=5$ items per campaign.}
    \label{tab:datasets}
    \setlength{\tabcolsep}{10pt}
    \begin{tabular}{@{}llrrll@{}}
        \toprule
        \textbf{Dataset} & \textbf{Source} & \textbf{Users} & \textbf{Items} & \textbf{Filter} & \textbf{[$\tau^{U}_{\min}, \tau^{U}_{\max}$]} \\
        \midrule
        Synthetic           & Generated   & 15\,000--40\,000 & 1\,500   & ---    & [0, 1\,500] \\
        IKEA                & Proprietary & 49\,962          & 23\,551  & 5-core & [200, 1\,000] \\
        Baby Products       & Amazon      & 150\,777         & 36\,002  & 5-core & [5\,000, 20\,000] \\
        Musical Instruments & Amazon      & 361\,313         & 149\,515 & 70\,\% & [3\,000, 10\,000] \\
        Interior Design     & Proprietary & 500\,000         & 121\,700 & ---    & [2\,000, 8\,000] \\
        \bottomrule
    \end{tabular}
\end{table*}

All methods are run 10 times and results are reported as mean $\pm$ 95\% confidence intervals. Biclustering is reproducible for a fixed \texttt{random\_state}, but not seed-independent. \texttt{SpectralCoclustering} is randomized in both the SVD and the $k$-means step, so it is repeated across 10 seeds like the others, with $\alpha$ re-selected within each repetition. Within each repetition we fit the full grid $\alpha \in \{2,3,4,5\}$ and report the value that maximizes average lift, so the reported intervals reflect variability in the selected model as well as in the fit.

The affinity matrix, $\mathbf{A}$, is precomputed and stored in memory throughout optimization. All experiments were run on a compute cluster of Intel Xeon E7-8860v4 nodes with 3072 GiB RAM per node. Runtimes are reported in Table~\ref{tab:runtime}.

\paragraph{Reproducibility.} Code, hyperparameter configurations, and the methods are available at \url{https://anonymous.4open.science/r/auto-targeting-3C48/}.

\paragraph{Dataset characteristics}
We evaluate on synthetic data (controlled, known structure), two Amazon Review benchmarks \cite{amazon} (reproducible, real-world), and proprietary IKEA and interior design datasets (practical deployment). Table~\ref{tab:datasets} summarizes the dataset characteristics.

\section{Results and Discussion}
\label{Section:results}

The objective of this evaluation was not to recover a global optimum, which is computationally infeasible at the scales we consider, but to identify methods that achieve high-quality campaign assignments under practical runtime constraints. We therefore assess performance as a trade-off between solution quality and computational efficiency.

\begin{table}[tb]
    \centering
    \caption{Results across all datasets. Within each dataset, joint methods are listed
    first and decoupled baselines follow after the gap. Stochastic methods
    report mean $\pm$ 95\% confidence intervals over 10 runs. Simulated Annealing is run once with a fixed random seed due to its high computational cost. The best values per
    dataset are shown in \textbf{bold}.}
    \label{tab:all_results}
    \footnotesize
    \setlength{\tabcolsep}{8pt}
    \begin{tabular}{@{}ll r@{\,$\pm$\,}l r@{\,$\pm$\,}l r@{\,$\pm$\,}l@{}}
        \toprule
        \textbf{Data} & \textbf{Method}
         & \multicolumn{2}{c}{\textbf{Lift}\,$\uparrow$}
         & \multicolumn{2}{c}{\textbf{Gini}\,$\downarrow$}
         & \multicolumn{2}{c}{\textbf{Qual.}\,$\uparrow$} \\
        \midrule
        \multirow{9}{*}{IKEA}
         & Greedy           & 2.22 & 0.04 & 0.03 & 0.01 & 11.2 & 0.2 \\
         & Biclustering     & \dt{\textbf{2.31\ci{0.00}}} & \dt{\textbf{0.02\ci{0.00}}} & \dt{\textbf{11.6\ci{0.03}}} \\
         & Thompson         & 2.23 & 0.03 & 0.03 & 0.01 & 11.3 & 0.2 \\
         & UCB1             & 2.24 & 0.03 & 0.03 & 0.01 & 11.3 & 0.2 \\
         & Sim.\ Ann.       & \dt{1.11} & \dt{0.15} & \dt{5.6} \\
        \addlinespace[0.5em]
         & \texttt{random}  & 1.10 & 0.02 & \textbf{0.02} & \textbf{0.01} & 5.53 & 0.11 \\
         & \texttt{kmeans}  & 1.27 & 0.01 & 0.05 & 0.00 & 6.42 & 0.02 \\
         & \texttt{top\_dim} & 1.98 & 0.00 & 0.05 & 0.00 & 9.96 & 0.00 \\
         & \texttt{popular} & 1.95 & 0.00 & 0.04 & 0.00 & 9.82 & 0.00 \\
        \midrule
        \multirow{9}{*}{\makecell[l]{Baby\\Prod.}}
         & Greedy           & 170 & 1 & 0.37 & 0.01 & 1513 & 10 \\
         & Biclustering     & \textbf{173} & \textbf{3} & \textbf{0.18} & \textbf{0.02} & \textbf{1543} & \textbf{29} \\
         & Thompson         & 168 & 2 & 0.36 & 0.01 & 1497 & 17 \\
         & UCB1             & \textbf{173} & \textbf{1} & 0.33 & 0.01 & 1541 & 9 \\
         & Sim.\ Ann.       & \dt{93} & \dt{0.44} & \dt{832} \\
        \addlinespace[0.5em]
         & \texttt{random}  & 1.10 & 0.23 & 0.19 & 0.10 & 9.83 & 2.09 \\
         & \texttt{kmeans}  & 1.55 & 0.00 & 0.38 & 0.00 & 13.83 & 0.01 \\
         & \texttt{top\_dim} & 17.94 & 0.00 & 0.25 & 0.00 & 160.1 & 0.0 \\
         & \texttt{popular} & 142.5 & 0.0 & 0.25 & 0.00 & 1272 & 0 \\
        \midrule
        \multirow{9}{*}{\makecell[l]{Musical\\Instr.}}
         & Greedy           & 91 & 12 & 0.45 & 0.04 & 777 & 102 \\
         & Biclustering     & \textbf{152} & \textbf{1} & \textbf{0.37} & \textbf{0.00} & \textbf{1300} & \textbf{6} \\
         & Thompson         & 89 & 9  & 0.43 & 0.03 & 762 & 73 \\
         & UCB1             & 88 & 11 & 0.51 & 0.04 & 756 & 96 \\
         & Sim.\ Ann.       & \dt{3} & \dt{0.69} & \dt{28} \\
        \addlinespace[0.5em]
         & \texttt{random}  & 5.69 & 1.01 & \textbf{0.35} & \textbf{0.07} & 49.26 & 8.74 \\
         & \texttt{kmeans}  & 4.11 & 0.00 & 0.49 & 0.00 & 35.59 & 0.00 \\
         & \texttt{top\_dim} & 14.74 & 0.00 & 0.44 & 0.00 & 127.7 & 0.0 \\
         & \texttt{popular} & 48.02 & 0.00 & 0.50 & 0.00 & 416.0 & 0.0 \\
        \midrule
        \multirow{9}{*}{\makecell[l]{Int.\\Design}}
         & Greedy           & 71 & 3 & 0.61 & 0.04 & 605 & 23 \\
         & Biclustering     & \textbf{176} & \textbf{8} & 0.55 & 0.02 & \textbf{1507} & \textbf{70} \\
         & Thompson         & 71 & 7 & 0.58 & 0.06 & 611 & 60 \\
         & UCB1             & 72 & 5 & 0.64 & 0.05 & 612 & 42 \\
         & Sim.\ Ann.       & \dt{2} & \dt{0.64} & \dt{4} \\
        \addlinespace[0.5em]
         & \texttt{random}  & 8.01 & 1.20 & 0.43 & 0.04 & 68.3 & 10.2 \\
         & \texttt{kmeans}  & 1.31 & 0.00 & \textbf{0.11} & \textbf{0.00} & 11.2 & 0.0 \\
         & \texttt{top\_dim} & 17.01 & 0.00 & 0.38 & 0.00 & 145 & 0 \\
         & \texttt{popular} & 36.82 & 0.00 & 0.34 & 0.00 & 314 & 0 \\
        \bottomrule
    \end{tabular}
\end{table}

\begin{table}[tb]
    \centering
    \caption{Runtime in minutes (except Simulated Annealing which is reported in hours). Stochastic methods report mean results with 95\% confidence intervals, computed over 10 runs.}
    \label{tab:runtime}
    \footnotesize
    \setlength{\tabcolsep}{6pt}
    \begin{tabular}{@{}l *{4}{r@{}l}@{}}
        \toprule
        \textbf{Method}
         & \multicolumn{2}{c}{\textbf{IKEA}} & \multicolumn{2}{c}{\textbf{Baby Prod.}}
         & \multicolumn{2}{c}{\textbf{Musical Instr.}} & \multicolumn{2}{c}{\textbf{Int. Design}} \\
        \cmidrule(lr){2-3}\cmidrule(lr){4-5}\cmidrule(lr){6-7}\cmidrule(lr){8-9}
        Greedy      & 7.4 & \ci{2.3} & 56.2 & \ci{0.1} & 74.3 & \ci{15.7} & 63.7 & \ci{10.9} \\
        Biclustering  & \textbf{1.1} & \textbf{\ci{0.1}} & \textbf{5.2} & \textbf{\ci{0.1}} & 62.3 & \ci{2.2} & 71.1 &\ci{3.8} \\
        Thompson    & 4.5 & \ci{0.9} & 7.9 & \ci{1.6} & 51.9 & \ci{5.5} & 68.6 & \ci{9.1} \\
        UCB1        & 1.6 & \ci{0.1} & 5.8 & \ci{0.4}
                    & \textbf{45.0} & \ci{\textbf{3.0}}
                    & \textbf{54.6} & \ci{\textbf{4.0}} \\
        Sim.\ Ann.
                      & \multicolumn{2}{c}{6.5 (h)}
                      & \multicolumn{2}{c}{37.4 (h)}
                      & \multicolumn{2}{c}{95.3 (h)}
                      & \multicolumn{2}{c}{98.9 (h)} \\
        \bottomrule
    \end{tabular}
\end{table}

The decoupled baselines determine what is reachable without joint optimization, and the gap to the joint methods measures what the coupling adds. The best decoupled method reaches a lift of 1.98 on IKEA and 142.5 on Amazon Baby Products, against 2.31 and 173 for the best joint method, gaps of 17\% and 21\%. On the two largest datasets the gap is even wider. The best decoupled pipeline reaches 48.0 on Amazon Musical Instruments and 36.8 on Interior Design, against 152 and 176, factors of 3.2 and 4.8. Because the user assignment is solved optimally in every baseline, these gaps come from the item selection alone.
Comparing the decoupled methods shows what the objective rewards in an item set. Selecting individually strong items (\texttt{popular}) beats clustering the item embeddings (\texttt{kmeans}) on every dataset, by factors between 1.5 and 92. On the two largest datasets \texttt{kmeans} performs worse than selecting items at random, reaching a lift of 4.11 against 5.69 on Musical Instruments and 1.31 against 8.01 on Interior Design. Items that lie close together does not need to be items that any one audience wants.

Among the joint methods, biclustering achieves the highest lift and quality on every dataset (Table~\ref{tab:all_results}). Greedy and bandit optimization, which score individual users and items, recover roughly
twice the best decoupled result on the two largest datasets, while biclustering recovers
three to five times as much. In terms of runtime (Table~\ref{tab:runtime}), the bandit-based methods are generally the most efficient on larger datasets, while greedy optimization is slower due to the repeated local search steps. Biclustering exhibits a more variable profile where it is highly efficient on smaller datasets, but becomes computationally expensive as the dataset sizes increase. For example, biclustering completed in 72 seconds on IKEA but required over 3,900 seconds on the interior design dataset, illustrating that its runtime scales steeply with the size of the affinity matrix. Appendix~\ref{app:convergence_dynamics} reports the convergence dynamics of each method.

The relative performance differences depend on the structure of the underlying interaction data. For Musical Instruments, the less restrictive 70\% filtering retains a broader and less structured interaction matrix, while the 5-core filtering used for Baby Products keeps only users and items with at least five interactions, producing a denser matrix with more shared interaction patterns.  In this setting, the underlying structure is easier to recover, and the performance gap between methods is smaller. This suggests that when the underlying structure is easier to recover, simpler optimization strategies can perform competitively. 

On the IKEA dataset, all the joint methods follow similar trends, with biclustering achieving the best overall performance. However, absolute scores are lower compared to the Amazon datasets, reflecting weaker affinities in the interaction data and less pronounced cluster structure.
For the interior design dataset, biclustering substantially outperforms the other methods, achieving more than double the lift and quality. This indicates that capturing joint user--item structure is particularly important in large-scale commercial settings. At the same time, all joint methods exhibit relatively high Gini coefficients, suggesting that a small subset of users contributes disproportionately to the overall performance. Biclustering mitigates this effect slightly, producing more balanced assignments.

SA performs substantially worse than the proposed methods across all datasets (\Cref{tab:all_results}), while also incurring substantially higher computational cost (\Cref{tab:runtime}). A likely explanation is that the local, single-entity moves used in SA are not well suited to the high-dimensional and combinatorial structure of the user--item assignment problem. As a result, the method struggles to make meaningful progress within a reasonable runtime, particularly on large datasets.

\subsection{Synthetic Data}

\begin{figure}[tb]
    \centering
  \includegraphics[width=\linewidth]{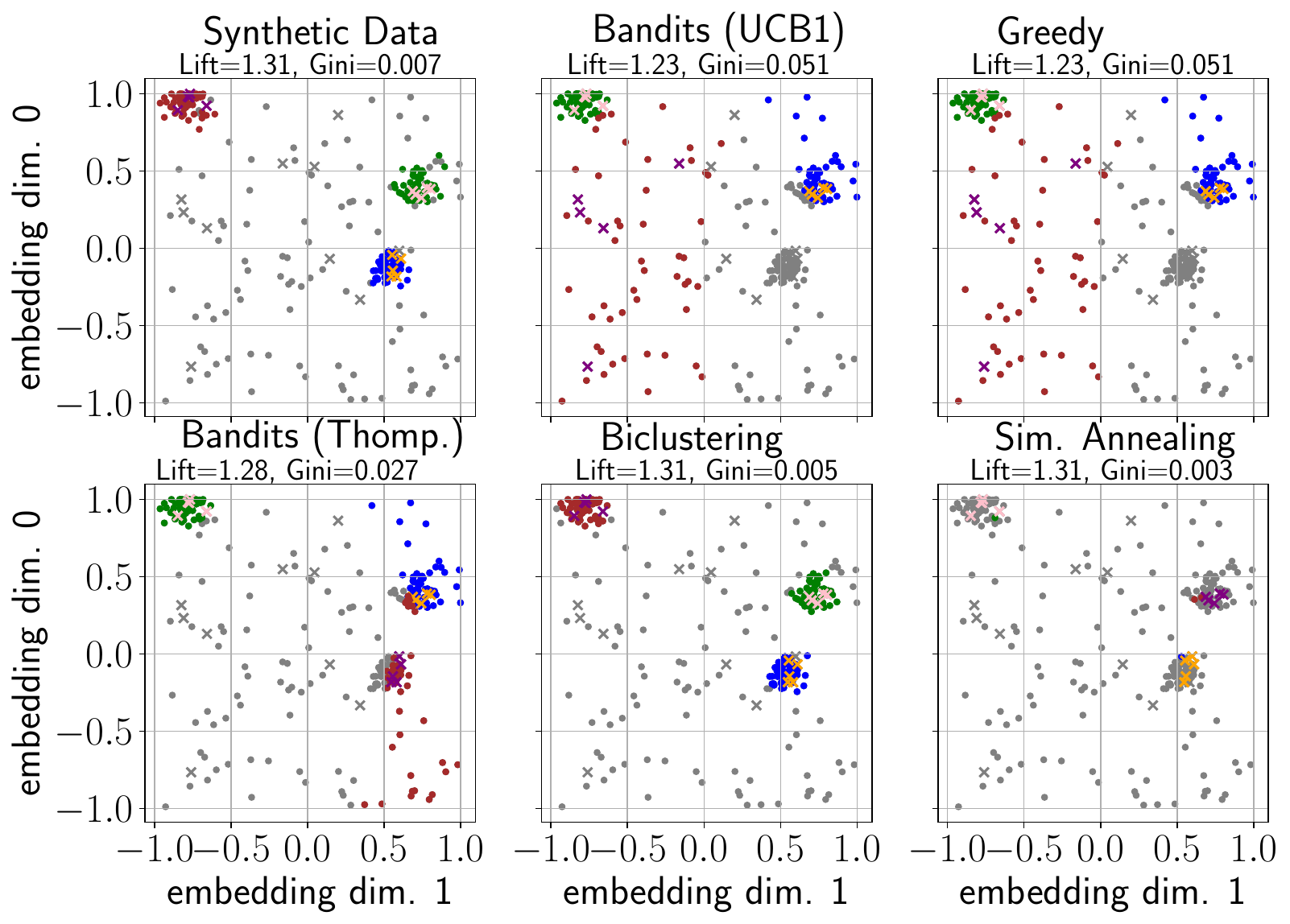}
    \caption{Small-scale 2-d synthetic dataset with 25 items and 250 users including 3 campaigns with 5 items and 50 users. Random baseline campaigns are compared across the given methods.}
    \label{fig:synth_small}
\end{figure}

\begin{figure}[t]
    \centering
    \includegraphics[width=\linewidth]{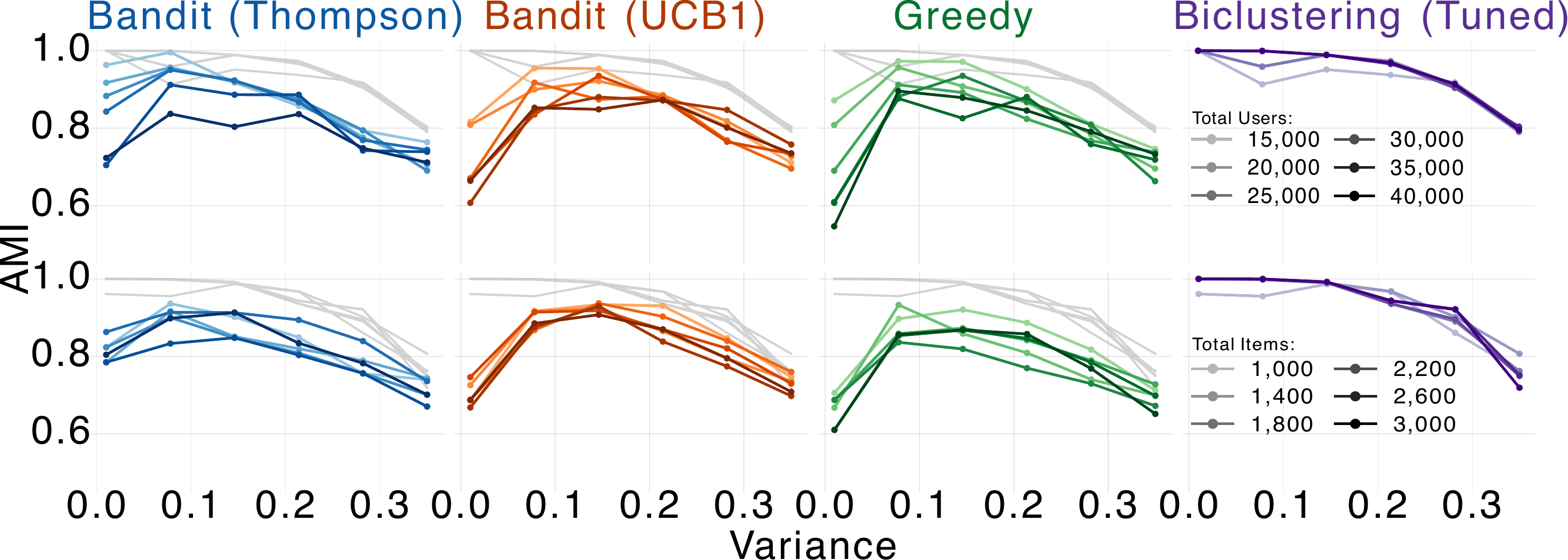}
    \caption{Recovery of the planted campaign structure on synthetic data, measured by AMI against the ground truth, as problem size vary. Campaigns contain at most 1,500 users and 5 items. Top row: the total number of users varies, with items fixed at 1,500. Bottom row: the total number of items varies, with users fixed at 30,000. Biclustering is repeated in grey in the other panels for reference.}
    \label{fig:synthetic_variance}
\end{figure}

\begin{figure*}[t]
    \centering
    \includegraphics[width=\linewidth]{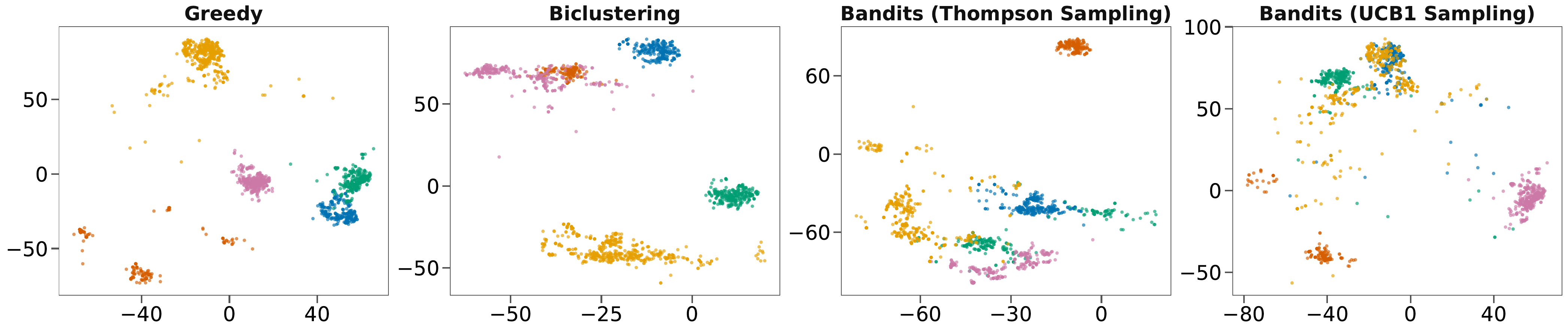}
    \caption{A t-SNE representation of user and item embeddings for the Amazon Musical Instruments dataset, colored by campaign assignment. Each panel shows the result of a different method.}
    \label{fig:t-sne_instruments}
\end{figure*}

To evaluate the campaign assignment algorithms under controlled conditions, we generate a synthetic dataset with a known underlying campaign structure. The $D$-dimensional user and item embeddings are drawn from a multivariate Gaussian centered at a campaign-specific vector. A small variance ensures that entities within the same campaign are closely aligned, leading to high affinity within campaigns.
To simulate realistic sparsity and background noise, additional items and users are drawn randomly from a uniform distribution. 
Because the planted campaigns are known, we can compare assignments against them directly, using adjusted mutual information (AMI)~\cite{ami}. The added noise can itself form a better campaign than the planted one, so the planted campaigns are only an approximation of the best achievable.
Notably, for the small-case synthetic data in \Cref{fig:synth_small}, biclustering found those better campaigns.
The Gini score for biclustering improved compared to the initial campaigns.

\Cref{fig:synthetic_variance} presents an evaluation of the methods' performances as the problem becomes progressively more difficult by increasing the cluster variance and the number of users and items. 
As exemplified in \Cref{fig:synth_small}, doubling the variance would lead to overlaps between the green and blue campaigns.
A higher variance thus reduces the separation between campaigns, while additional users and items introduce noise, making the underlying structure more difficult to recover.
%
Biclustering recovers the planted structure almost exactly up to a variance of about 0.2, regardless of the number of users or items, and only then declines towards the other methods. The other three are non-monotone: they perform worst when campaigns are separated, peak at a variance of roughly 0.1, and decline thereafter.  
This suggests that overlaps between campaigns (controlled by variance) is more detrimental than increasing the number of background users and items, even as the problem scales in both dimensions.

\subsection{Analysis}\label{sec:analysis}

To understand why biclustering consistently outperforms the other methods, we examine the structure of the embedding space, the qualitative coherence of the top campaign, and the distributional properties of the affinity matrix for the Amazon datasets. 

\paragraph{Campaign structure in embedding space.}
We project user and item embeddings jointly into two dimensions using t-SNE \cite{maaten2008visualizing}, after PCA pre-reduction to 10 dimensions; t-SNE is preferred over a linear projection here because it preserves local neighborhood structure, making cluster separation visually interpretable. \Cref{fig:t-sne_instruments} shows the resulting projection for Amazon Musical Instruments, colored by campaign assignment (we randomly subsample 15,000 users and 8,000 items for ease of comparison; Amazon Baby
Products shows the same pattern, Appendix~\ref{app:campaign_exploration}). Under this projection, biclustering produces the most compact and well-separated clusters,
consistent with its quantitative performance (\Cref{tab:all_results}), while the greedy and bandit methods yield more diffuse groupings with greater inter-campaign overlap, particularly for Thompson Sampling. As t-SNE is a qualitative visualization, the quantitative results in \Cref{tab:all_results} remain the more reliable basis for comparison.

\begin{figure}[t!]
    \centering
    \includegraphics[width=\linewidth]{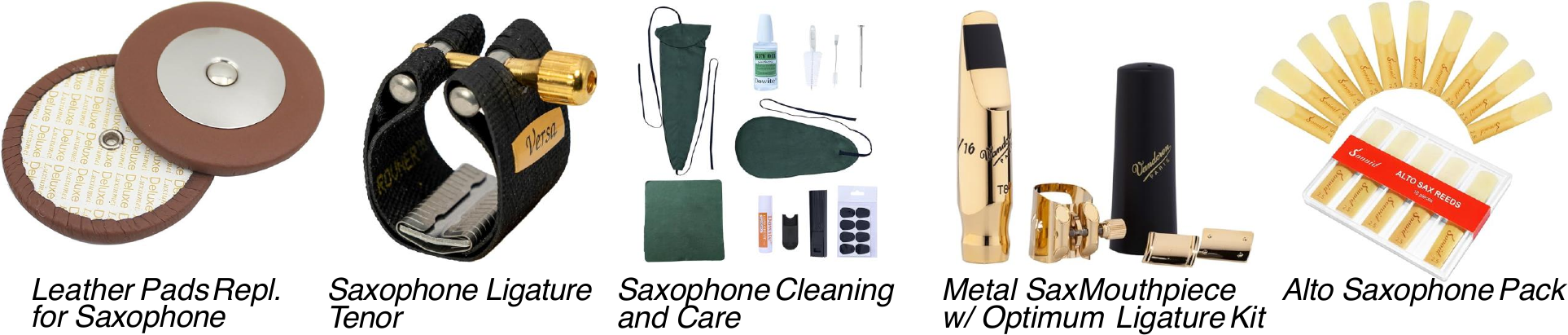}
    \caption{Qualitative analysis of campaigns produced by biclustering on the Amazon Musical Instruments dataset. The row shows the five items assigned to a single campaign, with shortened product titles and representative images.}
    \label{fig:campaign_instruments}
\end{figure}

\paragraph{Qualitative campaign coherence.}
Figure \ref{fig:campaign_instruments} shows the five items assigned to the
highest-quality campaign based on metric performance for the Amazon Musical Instruments dataset. Although the items span different product sub-categories, such as cleaning supplies, mouthpiece accessories, and replacement parts, they all serve the common goal of owning and maintaining a saxophone. Biclustering therefore recovers campaigns aligned with latent user intent rather than surface-level product taxonomy, an important property for deployment in real-world marketing systems where campaign coherence is crucial. Amazon Baby Products shows the same pattern (Appendix~\ref{app:campaign_exploration}).

\begin{figure}[tb]
    \centering
    \includegraphics[width=\linewidth]{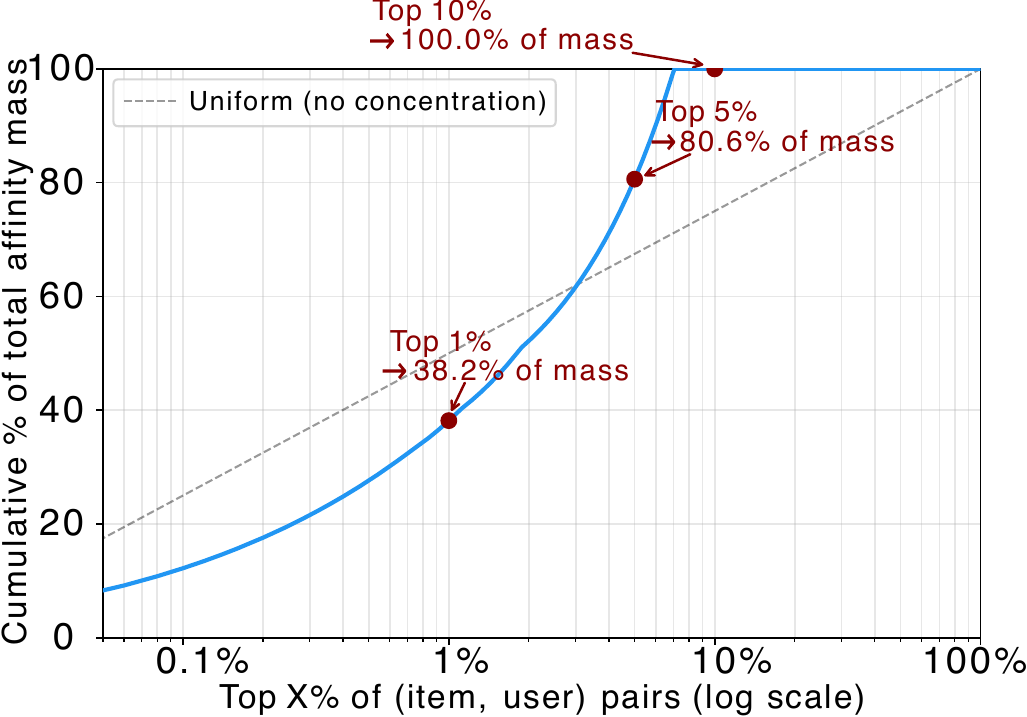}
    \caption{Affinity mass concentration across (item, user) pairs.}
    \label{fig:affinity_conc}
\end{figure}

\begin{figure}[t!]
    \centering
    \includegraphics[width=\linewidth]{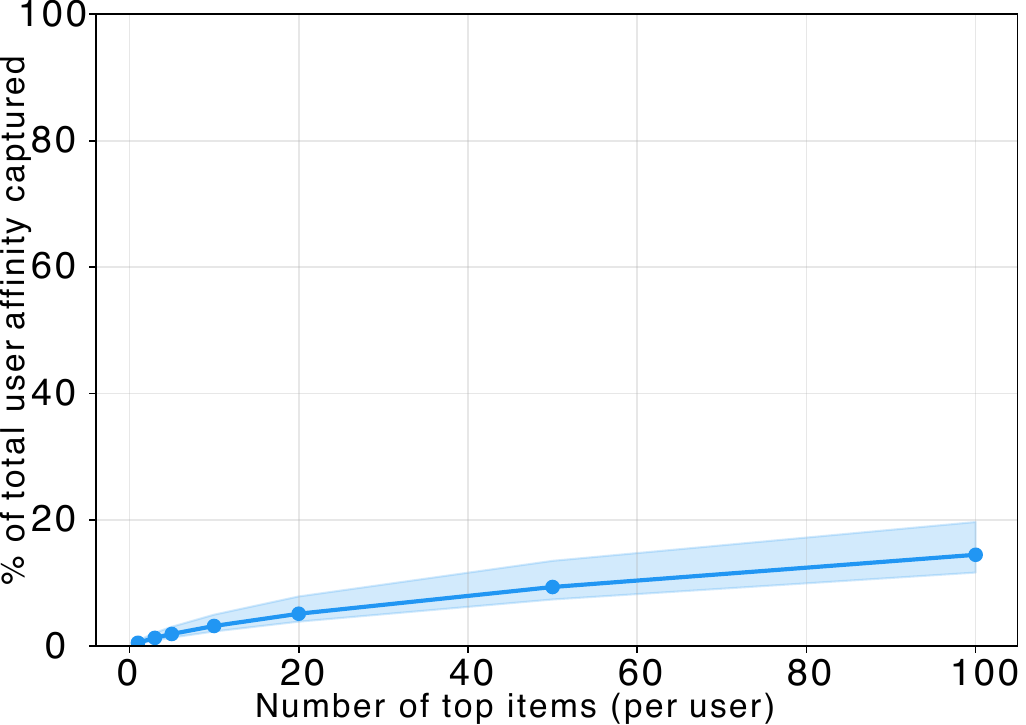}
    \caption{Per-user affinity concentration: median fraction of total user affinity captured by top-k items (shaded band = IQR), computed over an $8{,}000$-item $\times$ $15{,}000$-user affinity submatrix ($120$M entries) from Amazon Musical Instruments.}
    \label{fig:top_items_aff}
\end{figure}

\paragraph{Affinity concentration.}
The structure of the affinity matrix helps to further explain the results.
As seen in the Amazon Musical Instruments dataset in  Figure~\ref{fig:affinity_conc}, the distribution of affinity values is heavily right-skewed: the top $1\%$ of $(i,u)$ pairs account for $38\%$ of total affinity mass, and the top $10\%$ account for essentially all of it. At the same time, per-user preferences are diffuse. For the
median user, the single highest-affinity item captures only $0.5\%$ of their total affinity, and the top 100 items capture just $14.5\%$
(Figure~\ref{fig:top_items_aff}). Because the signal is globally concentrated in a small fraction of pairs but no single pair dominates for any given user, methods that operate on individual entities, such as greedy assignment, lack
a strong signal to act on. Biclustering, by contrast, aggregates many weak per-user affinities across a block of users and items simultaneously, recovering coherent high-affinity regions that would not be apparent from a single entry alone. Across the $K=5$ campaigns, $92.4\%$ of the cells selected by biclustering rank in the global top $1\%$ of affinity values, while covering only $0.0038\%$ of the matrix, with campaign-level lift over the global mean ranging from $60\times$ to
$171\times$, confirming that biclustering reliably identifies high-signal submatrices rather than noise.

\subsection{Limitations and Future Work}

This work is motivated by campaign delivery on commercial Web platforms, where automated assignment could improve the relevance and efficiency of promotional targeting. While this has potential benefits for both retailers and users, it also raises concerns around privacy, as the framework relies on learned user embeddings derived from behavioral data logged by the platform. Because campaigns are disjoint and capacity-bounded, some users necessarily receive a less well-matched campaign than an unconstrained assignment would give them. Deployment in practice should be accompanied by appropriate transparency about how audiences are formed and by fairness considerations, particularly to avoid reinforcing existing disparities in how different user groups are targeted. 

While the proposed framework provides a flexible approach to joint user--item campaign assignment, several limitations remain. The methods are heuristic and do not provide guarantees on solution quality, and the greedy and bandit-based methods may converge to local optima. Scalability remains a challenge for very large datasets, as biclustering becomes computationally expensive as the affinity matrix grows. The formulation also assumes a static setting with fixed user--item affinities, whereas in practice user preferences and item availability evolve over time. We also take the embedding space as given, so our results characterize allocation performance under one particular embedding geometry rather than its sensitivity to the choice of embedding model. Future work should examine how auto-targeting performs under different latent spaces. Finally, we consider a simplified constraint structure; incorporating overlapping audiences, budget limits, or fairness requirements across user groups remains an open challenge.

\section{Conclusions}

We formalized campaign assignment as auto-targeting: assigning users and items to disjoint, size-bounded campaigns that maximize within-campaign affinity, with neither the user nor the item side fixed in advance. Joint optimization pays off. Against decoupled baselines whose user assignment is solved exactly, the best joint method wins on every dataset, by 17\% and 21\% on the two smaller, denser datasets, and by 220\% and 380\% on the two largest. The joint problem is also tractable. All three joint methods solve the largest datasets well in about an hour, and the bandit methods are faster than biclustering where the data are largest. Biclustering, which works on blocks of the affinity matrix, recovers the most coherent campaigns and matches or exceeds the quality of the other joint methods on every dataset, with the greatest margin on the sparsest data. The best results always come from selecting items and audiences jointly.

\begin{acks}
We thank Theodor Jonsson and Christian Persson at Siftlab AB for helpful discussions and help with generating the embeddings.
M.L.\ and N.G.\ were supported by the Wallenberg AI, Autonomous Systems and Software Program (WASP), funded by the Knut and Alice Wallenberg Foundation.
J.P.\ acknowledges funding by the German Ministry of Research, Technology and Space (BMFTR; no.~ 16IS24072E; COMFORT).
M.R.\ was supported by the Swedish Research Council (no.~2023-03705).
The computations were enabled by resources provided by the Swedish National Infrastructure for Computing at the High Performance Computer Center North (HPC2N) in Umeå, Sweden, partially funded by the Swedish Research Council ( no.~2018-05973), and by the National Academic Infrastructure for Supercomputing in Sweden (NAISS), partially funded by the Swedish Research Council (no.~2022-06725).
\end{acks}

\appendix

\section{Implementation Details}\label{app:implementation}

This appendix provides implementation details that complement the method descriptions in Section~\ref{sec:approaches}. All algorithms use the same affinity definition,
$\mathbf{A}_{iu} = \exp(\langle \mathbf{V}_i, \mathbf{U}_u \rangle)$.

\subsection{Score Aggregation}\label{app:score_aggregation}

The affinity of entity $e$ to campaign $k$ is computed as a dense matrix product. Letting $\mathbf{Z}^{I} \in \{0,1\}^{N_i \times K}$ and
$\mathbf{Z}^{U} \in \{0,1\}^{N_u \times K}$ be the current item- and user-assignment
indicator matrices, we form
\begin{equation}
    \mathbf{S}^{U} = \mathbf{A}^{\top}\mathbf{Z}^{I} \in \mathbb{R}^{N_u \times K},
    \qquad
    \mathbf{S}^{I} = \mathbf{A}\,\mathbf{Z}^{U} \in \mathbb{R}^{N_i \times K},
\end{equation}
where column $k$ of $\mathbf{S}^{U}$ (resp. $\mathbf{S}^{I}$) gives $s^u_k$ (resp. $s^i_k$)
for every user (resp. item). The full affinity matrix $\mathbf{A}$ is pre-computed before the optimization begins.

\subsection{Greedy Local Search: Allocation and Swap Details}

\emph{Heap-greedy allocation.} Entities are processed with a min-heap (negated scores).
Each entity is initialized with its best feasible campaign and, once assigned, is
re-enqueued with its next-best option. Stale heap entries, where the stored score differs
from the current score by more than $10^{-6}$, are discarded. Campaign capacities are tracked with a count vector. Capacity checks are only required when an entity first enters a campaign, since swaps between entities within the same campaign leave capacities unchanged.

\emph{Sliding-window local search.} At the start of each pass, entity indices are shuffled. For an entity $e_1$ at position $p$, positions $p+1, \dots, p+W \pmod{N_e}$ are considered as swap candidates, where $N_e$ is the number of entities being assigned and the window size is set to $W = 128$. A swap $(e_1, k_1) \leftrightarrow (e_2, k_2)$ is accepted when the entities belong to different campaigns, the joint score improves,
\begin{equation*}
    s_{k_2}(e_1) + s_{k_1}(e_2) > s_{k_1}(e_1) + s_{k_2}(e_2),
\end{equation*}
and capacity constraints remain satisfied. Each local search phase runs for at most $P = 10$ passes, with early stopping triggered after three consecutive passes with no accepted swaps.

\emph{Outer loop.} The algorithm alternates between user and item assignment for at most
$T = 10$ outer iterations. Early stopping is applied when the improvement in the objective
falls below $\epsilon = 10^{-4}$ for three consecutive iterations. The best solution
encountered across all iterations is retained.

\subsection{Bandit Optimization: Algorithmic Details}

The bandit approach follows the same alternating structure as the greedy method but replaces heap-greedy allocation with bandit-augmented scoring. We run the optimization for up to $T = 50$ outer iterations, with early stopping triggered after ten consecutive iterations whose improvement falls below $\epsilon = 10^{-4}$.

\emph{UCB1 bonus scale.} The mean score magnitude used in the UCB1 bonus is defined as
\begin{equation*}
    \hat{s}_k = \frac{1}{N_e} \sum_e |s_{ek}| + 10^{-8},
\end{equation*}
which ensures that the exploration bonus is commensurate with the reward magnitude. The
exploration coefficient $c$ is treated as a hyperparameter and selected via a sweep over
$c \in \{0.1, 0.5, 1.0, 2.0, 5.0, 10.0\}$, choosing the value that yields the highest final quality.

\emph{Thompson Sampling noise.} For Thompson Sampling, a noise vector
$\boldsymbol{\eta}_k \in \mathbb{R}^{N_e}$ is drawn from
\begin{equation*}
    \mathcal{N}(0, \sigma_k \hat{s}_k), \quad \text{with} \quad
    \sigma_k = (n_k + 1)^{-1/2},
\end{equation*}
where the $+1$ term ensures a well-defined variance when campaign $k$ has received no
assignments ($n_k = 0$). The count $n_k$ is incremented after each assignment to campaign
$k$.

\emph{Greedy assignment with bandit scores.} Entities are inserted into a max-heap keyed by $\tilde{s}_{ek^*}$, where $k^* = \arg\max_k \tilde{s}_{ek}$. When a campaign reaches capacity, its augmented score is set to $-\infty$ and the entity is re-enqueued with its next-best option. Bandit statistics $(n_k, \sum r_k, \sum r_k^2)$ are updated using the unaugmented reward, $s_{ek}$.

\emph{Non-stationarity.} To account for non-stationarity, we optionally apply exponential
decay with window parameter $w$, updating statistics each round by a factor $(1 - 1/w)$.

\emph{Regret decomposition.} The cumulative regret satisfies
$R(T) \leq R_{\mathrm{user}}(T)$, since item reassignment can only improve or maintain the
score obtained after user assignment. The gap quantifies how often item updates correct
suboptimal user assignments.

\subsection{Constrained Spectral Biclustering: Algorithm Details}\label{app:spectral}

We implement spectral co-clustering \urlstyle{tt} using \path{sklearn.cluster.SpectralCoclustering} with \texttt{random\_state=0}. Given a target of $K$ campaigns, we generate $K' = \alpha K$ candidate biclusters, where
$\alpha$ is selected via a sweep over $\alpha \in \{2, 3, 4, 5\}$, choosing the value that yields the highest final quality after constraint enforcement.

Each candidate bicluster is pruned to satisfy minimum and maximum size constraints by selecting the rows and columns with highest average affinity to the opposing side. Biclusters that do not satisfy minimum size constraints are discarded.

The remaining candidates are ranked by their density, and the top $K$ non-overlapping biclusters are selected as campaigns. Unlike the greedy and bandit approaches, biclustering is a single-pass procedure ($T = 1$) and does not iteratively optimize the objective in Eq.~\ref{eq:problem}, as it is based on a normalized-cut relaxation.

\subsection{Simulated Annealing: Algorithmic Details}\label{app:SA}

Our approach employs a variant of simulated annealing tailored for the discrete nature of the targeting problem. The algorithm maintains temperature-dependent exploration via probabilistic acceptance of moves.

\paragraph{State Representation and Neighborhood}

The state of a campaign is represented as a tuple $(I, U)$ of item and user indices. The neighborhood is defined through three mutation operations:
\begin{itemize}
    \item \textbf{Add}: Randomly select an available item or user and add it to the current set.
    \item \textbf{Remove}: Randomly remove an item or user from the current set (if constraints allow).
    \item \textbf{Swap}: Remove one item/user and add a different available one.
\end{itemize}
The algorithm randomly selects add/remove/swap with equal probability (subject to feasibility), ensuring diverse exploration of the solution space.

\paragraph{Acceptance Criterion}

At iteration $t$, given temperature $T_t$ and current score $s_t$, a move to state with score $s'_t$ is accepted if:
\begin{align}
s'_t > s_t \quad \text{or} \quad \mathcal{U}(0,1) < \exp\left(\frac{\Delta s}{T_t}\right),
\quad \text{where} \quad \Delta s = s'_t - s_t
\end{align}
This is the standard Metropolis criterion, enabling uphill moves with decreasing probability as temperature decreases.

\paragraph{Cooling Schedule and Early Stopping}

The temperature follows a geometric cooling schedule:
\begin{align}
T_t = T_0 \cdot \alpha^t
\end{align}
where $T_0 = 2.0$ is the initial temperature and $\alpha = 0.9999995$ is the cooling rate. The algorithm terminates when either:
\begin{enumerate}
    \item $T_t$ drops below $T_{\min} = 0.001$, or
    \item After $L_{\min} = 6 \cdot 10^6$ iterations, if relative score improvement over the last $P = 100{,}000$ iterations is below tolerance $\epsilon = 10^{-5}$.
\end{enumerate}

To enable early stopping with useful intermediate estimates while allowing selective refinement, we employ a two-phase optimization:

\paragraph{Phase 1: Quick Initial Pass}
Each campaign is optimized for $L_1 =$ \mbox{$1 \cdot 10^6$} iterations, providing preliminary solutions. This phase provides baseline performance estimates for all campaigns within reasonable wall-clock time.

\paragraph{Phase 2: Adaptive Refinement}
Campaigns are sorted by Phase 1 scores in descending order. Each campaign then undergoes refinement for an additional $L_2 = 5 \cdot 10^6$ iterations. Crucially, each campaign is refined \textit{independently} with access to the full original user/item pools, minus only the selections already made by previously-refined campaigns in Phase 2. This design ensures that high-performing campaigns are refined with maximal available resources, avoiding the degradation that occurs when resource pools are sequentially depleted.

\section{Convergence dynamics}
\label{app:convergence_dynamics}
\begin{figure}[t!]
    \centering
    \includegraphics[width=\linewidth]{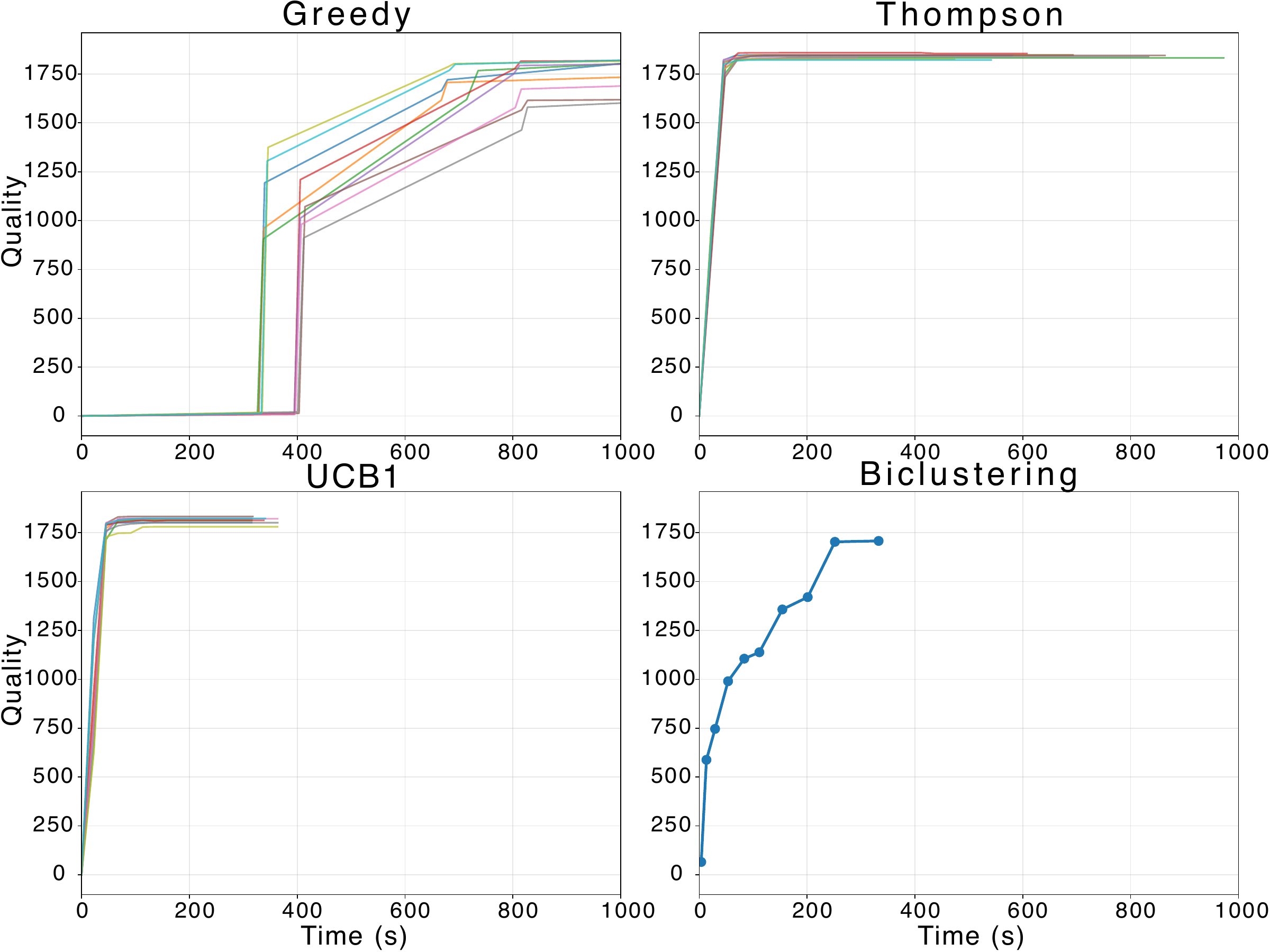}
    \caption{Quality progression of the Amazon baby products dataset over time.}
    \label{fig:progression_comp}
\end{figure}

Figure~\ref{fig:progression_comp} shows quality versus wall-clock time on
Amazon Baby Products. For each stochastic method we run 10 trials and record quality after each optimization step (an outer iteration for Greedy, a round
of assignments for the bandits); biclustering is deterministic and
non-iterative, so we instead subsample $10\%$ to $100\%$ of users and items and recompute the affinity matrix at each fraction. Thompson Sampling and UCB1 converge rapidly, reaching near-peak quality within 50--100 seconds with low variance across trials. Greedy is markedly slower. Biclustering follows a different pattern, rather than iterating toward convergence, its quality grows steadily with the fraction of users and items included, reaching its maximum on the full dataset at around 310 seconds.

\section{Campaign exploration}\label{app:campaign_exploration}
This section reports the Amazon Baby Products counterparts of the analyses in Section~\ref{sec:analysis}. 

\begin{figure*}[t]
    \centering
    \includegraphics[width=\linewidth]{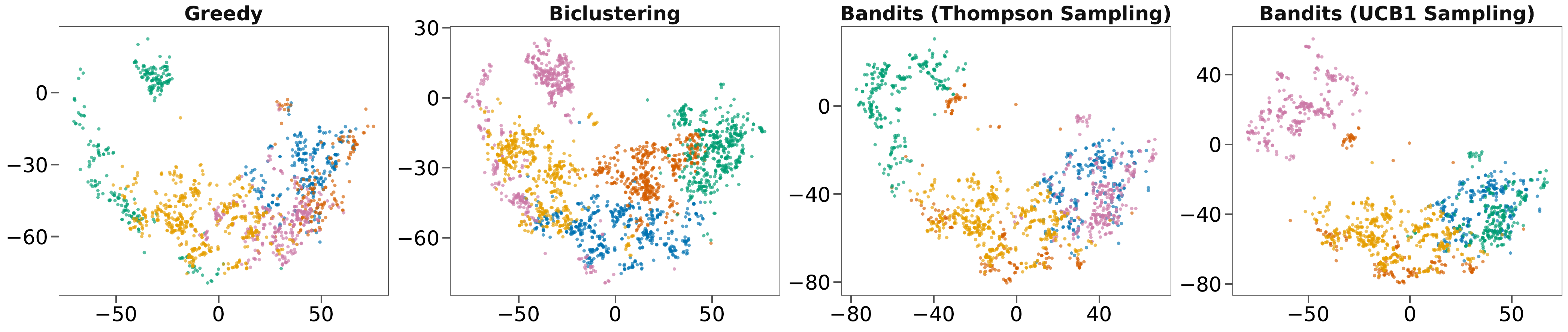}
    \caption{A t-SNE projection of user and item embeddings for the Amazon Baby Products dataset, colored by campaign assignment. Each panel shows the result of a different method.}
    \label{fig:t-sne_baby}
\end{figure*}

Figure~\ref{fig:t-sne_baby} projects the Baby Products user and item embeddings into two dimensions using the same procedure as for Musical Instruments: PCA pre-reduction to 10 dimensions followed by t-SNE. The pattern matches the one reported in the main text. Biclustering produces the most compact and best separated clusters, while the greedy and bandit methods yield more diffuse groupings with greater inter-campaign overlap, most visibly for Thompson Sampling.

\begin{figure}[t!]
    \centering
    \includegraphics[width=\linewidth]{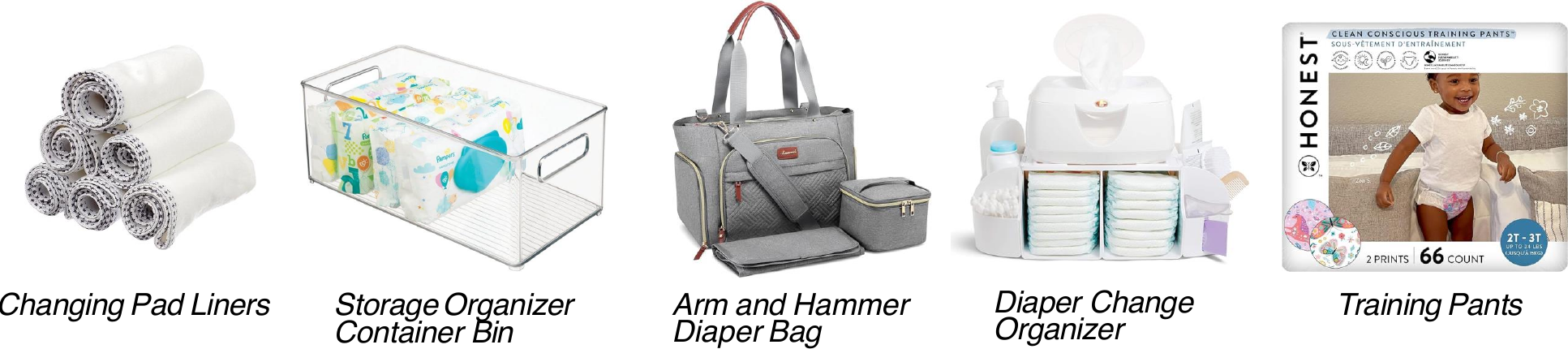}
    \caption{Qualitative analysis of campaigns produced by biclustering on the Amazon Baby Products dataset. The row shows the five items assigned to a single campaign, with shortened product titles and representative images.}
    \label{fig:campaign_baby}
\end{figure}

Figure~\ref{fig:campaign_baby} shows the five items assigned to the highest-quality Baby Products campaign under biclustering. The campaign spans cleaning pads, a diaper bag, a storage organizer, and training pants: items from different product categories unified by the goal of managing diaper needs. This is the same pattern as the saxophone campaign discussed in the main text, and supports the same conclusion, that biclustering recovers campaigns aligned with latent user intent rather than with surface-level product taxonomy.

\clearpage
\end{document}